\documentclass[pdflatex,sn-mathphys-num]{sn-jnl}

\usepackage{amsmath,amsfonts}
\usepackage{graphicx,wrapfig,lipsum}
\usepackage{graphicx}
\usepackage{textcomp}
\usepackage{xcolor}
\usepackage{color, colortbl}
\usepackage{url}      
\usepackage{hyperref}
\usepackage{comment}
\usepackage{float}
\usepackage{multirow}
\usepackage{amsthm}
\usepackage[title]{appendix}
\usepackage{manyfoot}%
\usepackage{booktabs}%
\usepackage{algpseudocode}%
\usepackage{listings}%
\usepackage[linesnumbered,ruled,vlined]{algorithm2e}
\usepackage{array}
\usepackage[caption=false,font=normalsize,labelfont=sf,textfont=sf]{subfig}
\usepackage{stfloats}
\usepackage{url}
\usepackage{verbatim}
\usepackage{hyperref}

\theoremstyle{thmstyleone}%
\theoremstyle{thmstyletwo}%

\theoremstyle{thmstylethree}%
\newcommand{\hl}[1]{\textcolor{black}{#1}}

\begin{document}

\title[Article Title]{CURENet: Combining Unified Representations for Efficient Chronic Disease Prediction}








\author[1,2]{Cong-Tinh Dao}\email{dctinh@ctu.edu.vn}
\author[1,2]{Nguyen Minh Thao Phan}\email{pnmthao@ctu.edu.vn}
\author[3]{Jun-En Ding}\email{jding17@stevens.edu}
\author[4]{Chenwei Wu}\email{chenweiw@umich.edu}
\author[5]{David Restrepo}\email{davidres@mit.edu}
\author[6]{Dongsheng Luo}\email{dluo@fiu.edu}
\author[3]{Fanyi Zhao}\email{fanyi.zhao@icloud.com}
\author[3]{Chun-Chieh Liao}\email{cliao9@stevens.edu}
\author[1]{Wen-Chih Peng}\email{wcpengcs@nycu.edu.tw}
\author[8]{Chi-Te Wang}\email{drwangct@gmail.com}
\author[8,9]{Pei-Fu Chen}\email{femh96949@mail.femh.org.tw}
\author[1]{Ling Chen}\email{ling.chen@nycu.edu.tw}
\author[7]{Xinglong Ju}\email{xinglongju@suu.edu}
\author[3]{Feng Liu}\email{fliu22@stevens.edu}
\author*[8]{Fang-Ming Hung}\email{philip@mail.femh.org.tw}

\affil[1]{National Yang Ming Chiao Tung University, Taiwan}
\affil[2]{Can Tho University, Vietnam}
\affil[3]{Stevens Institute of Technology, USA}
\affil[4]{University of Michigan, USA}
\affil[5]{Massachusetts Institute of Technology, USA}
\affil[6]{Florida International University, USA}
\affil[7]{Southern Utah University, USA}
\affil[8]{Far Eastern Memorial Hospital, Taiwan}
\affil[9]{Yuan Ze University, Taiwan}


\abstract{Electronic health records (EHRs) are designed to synthesize diverse data types, including unstructured clinical notes, structured lab tests, and time-series visit data. Physicians draw on these multimodal and temporal sources of EHR data to form a comprehensive view of a patient’s health, which is crucial for informed therapeutic decision-making. Yet, most predictive models fail to fully capture the interactions, redundancies, and temporal patterns across multiple data modalities, often focusing on a single data type or overlooking these complexities. In this paper, we present CURENet, a multimodal model (Combining Unified Representations for Efficient chronic disease prediction) that integrates unstructured clinical notes, lab tests, and patients' time-series data by utilizing large language models (LLMs) for clinical text processing and textual lab tests, as well as transformer encoders for longitudinal sequential visits. CURENet has been capable of capturing the intricate interaction between different forms of clinical data and creating a more reliable predictive model for chronic illnesses. We evaluated CURENet using the public MIMIC-III and private FEMH datasets, where it achieved over 94\% accuracy in predicting the top 10 chronic conditions in a multi-label framework. Our findings highlight the potential of multimodal EHR integration to enhance clinical decision-making and improve patient outcomes.}

\keywords{Large Language Model Fine-tuning, Transformer, Electronic Health Records, Multi-Disease Prediction}



\maketitle

\section{Introduction}

Noncommunicable diseases (NCDs), known as chronic illnesses, are typically long-term and are caused by a combination of genetic, physiological, environmental, and behavioral factors \cite{WHO_NCDs}. According to the World Health Organization (WHO), NCDs account for 41 million deaths each year, or 74\% of all deaths globally \cite{WHO_NCDs}. These conditions are widespread, impairing quality of life and placing a significant burden on healthcare systems. Chronic diseases remain among the leading causes of death \cite{fardet2022exclusive}. They are neither fully preventable nor curable, and their impact persists over time. However, early identification can help mitigate their impact. Consequently, the importance of automated chronic disease prediction in healthcare is increasingly apparent. Medical practitioners can take early action against the disease via computerized prediction algorithms that evaluate and identify high-risk individuals on the basis of various studies. These methods uncover hidden patterns in vast health datasets, helping clinicians focus on the most vulnerable patients and allocate resources more efficiently. Therefore, automated chronic disease prediction is essential for reducing the global burden of chronic diseases and advancing precision medicine.

\begin{figure}
    \centering
    \includegraphics[width=\textwidth]{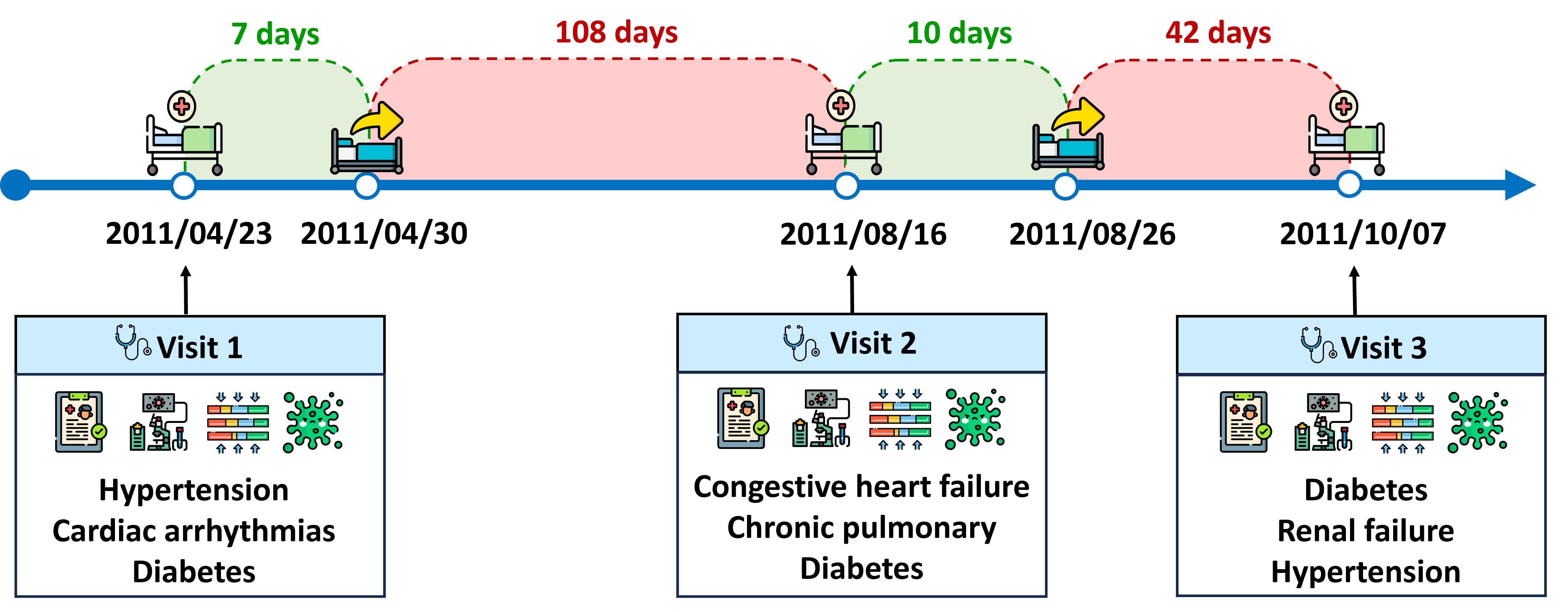}
    \caption{Example of EHR data with visits and Diagnoses with time durations and intervals.}
    \label{fig:intro}
\end{figure}

Electronic health records (EHRs), which combine unstructured clinical notes with structured tabular information, such as vital signs, diagnoses, and test findings, offered a rich, multimodal dataset. These many data sources, such as time series data from ongoing monitoring, provide a significant chance for disease prediction. The achievements of deep learning (DL) provide a great opportunity to improve the precision and efficacy of disease prediction. Despite the sophisticated models that can improve patient outcomes \cite{ding2024large, restrepo2024df, thao2024medfuse} and the abundance and variety of multimodal EHR data, it may be challenging to gather and integrate therapeutically valuable data from EHRs because of their redundancy and variety. Figure \ref{fig:intro} shows a patient's medical journey and the progression of their health over time with three hospital stays. The green areas represent the time duration, which is the length of hospital stay at each visit, whereas the red areas represent the time intervals between visits, which are computed from the discharge time of one visit to the admission of the next patient. Generally, diabetes was consistently diagnosed during all three visits, indicating that it is a chronic condition requiring ongoing care. It is the leading cause of a patient's general health issues and may be a precursor to more severe diseases. The intervals between visits -- 108 days between the first and second visits, and 42 days between the second and third visits -- show a concerning trend. The presence of multiple co-occurring conditions (multimorbidity) further complicates treatment, making this an ideal illustration of the complexities in EHR data.

The first primary problem in improving clinical predictions is as follows: \textit{Can we effectively extract and combine representations from diverse EHR modalities?} Current EHR modeling typically overlooks insights from unstructured clinical notes and lab tests, focusing instead on single data modalities \cite{luo2020hitanet,ma2017dipole}. This limited view can prevent models from capturing a complete picture of a patient's medical state. Traditionally, structured EHR data are modeled as numerical vectors, which ignore intricate relationships among variables \cite{choi2016retain,li2020behrt, zhang2020inprem,luo2020hitanet, ma2017dipole}. Lab tests, which are often high-dimensional and discrete, present additional challenges. Furthermore, temporal information in EHRs is frequently underused. While time series dependencies could yield important insights into a patient's longitudinal health status, many models rely on static representations instead.
Although large language models (LLMs) such as BERT \cite{devlin2018bert} have been effective with unstructured clinical notes and tasks such as few-shot predictions or medical question answering \cite{thirunavukarasu2023large}, research shows that LLMs still underperform on tabular and time series lab data \cite{grinsztajn2022tree, bellamy2023labrador, hegselmann2023tabllm, chung2024large}.

Another significant issue with time-stamped sequences is \textit{how predictive models can handle the complexity of several irregular visits?} Each patient's clinical timeline comprises a sequence of visits, and each visit may introduce new information, update previous diagnoses, or reveal emerging conditions. This dynamic nature makes multilabel disease prediction particularly complex.
Models must address missing data, variable visit intervals, and inconsistencies in follow-up.
Earlier models \cite{choi2016retain,choi2016doctor, kwon2018retainvis, ma2017dipole, baytas2017patient, bai2018interpretable} often failed to capture these nonlinear and multiscale relationships between visits. However, understanding these time relationships is vital since chronic conditions frequently progress and co-occur in complex, evolving ways \cite{wang2024recent}. Sophisticated modeling techniques are required to manage these irregularities and to leverage the temporal structure fully for accurate disease prediction.

In this study, we propose CURENet, a multimodal model \textbf{C}ombining \textbf{U}nified \textbf{R}epresentations for an \textbf{E}fficient chronic disease prediction that utilizes structured laboratory results, unstructured clinical notes, and time series visit data as different modalities for efficient chronic disease prediction. CURENet leverages fine-tuned LLMs to extract rich semantic features from text, whereas a transformer-based encoder captures irregular time sequences and disease progression patterns. This fusion enables a comprehensive understanding of a patient’s health status and supports improved diagnostic accuracy.

\hl{Our work contributes to the following key contributions:}



\begin{itemize}
    \item \hl{We proposed a novel cross-modal representation learning framework that effectively bridges unstructured clinical notes and temporal embeddings. This is achieved by fine-tuning the Medical-LLaMA3-8B model alongside a Time Series Transformer, outperforming traditional concatenation-based approaches.}

    \item \hl{We addressed the critical challenge of irregular visit patterns by incorporating both visit duration and inter-visit gaps into temporal modeling. This enables accurate predictions even when healthcare interactions are sparsely distributed -- a common occurrence in chronic illness treatment.}

    \item \hl{We achieved over 94\% accuracy in predicting the top 10 chronic conditions across two datasets (MIMIC-III and FEMH), showing consistent improvements (2-4\%) over state-of-the-art medical LLM baselines. This is supported by ablation studies, case analyses, and embedding visualizations that reveal better disease clustering than competing models do.}
\end{itemize}

\section{Related work}
Effective chronic disease prediction requires modeling complex relationships across multimodal clinical data and longitudinal health trajectories. Recent research has focused on integrating unstructured clinical notes, structured lab results, and temporal visit patterns. \hl{However, many models still fail to capture their interdependence. For instance, these existing models often process unstructured clinical notes and structured data in isolation, overlooking how textual information may contextualize or explain trends in lab results or vital signs. Consequently, they miss the interplay between narrative descriptions and quantitative measures. Furthermore, temporal dynamics are frequently modeled only on structured data, ignoring the implicit timing and progression cues in clinical notes, which limits the model’s ability to form holistic representations of patient trajectories.}

\subsection{Multimodal EHR Data for Disease Prediction}
Recent medical advancements have increasingly leveraged medical texts. When standardized codes fail to fully capture clinical symptoms or additional evidence is needed for diagnosis, unstructured text becomes valuable. For example, Kim et al. \cite{chen2015convolutional} developed a convolutional attention network to extract document-level representations. Other studies have employed graph neural networks (GNNs) to capture text-based entity relationships via attention processes \cite{chen2019deep} or to represent structured Electronic Medical Records (EMRs) as hierarchical graphs \cite{wu2021counterfactual}.
LLMs have shown strong performance in clinical tasks \cite{lu2024large, singhal2023large, peng2024depth, liu2023medical}, including natural language inference, according to the medical literature \cite{luo2022biogpt, bolton2024biomedlm}.

However, EHR data often combine unstructured notes with structured components such as lab results, imaging, and administrative records. Integrating these heterogeneous modalities remains a technical challenge. While data fusion models such as MedFuse \cite{thao2024medfuse} and DF-DM \cite{restrepo2024df} improve performance by unifying laboratory tests and clinical text with temporal signals, they often apply straightforward fusion methods that may not fully capture multimodal dependencies. LLMs are still evolving in their ability to handle unstructured data \cite{thirunavukarasu2023large, lu2024large}, but many struggle with tabular data such as lab results \cite{grinsztajn2022tree, bellamy2023labrador, hegselmann2023tabllm, chung2024large}.


Overall, previous studies often focus on a single modality \cite{miotto2016deep, cheng2016risk, choi2017gram, suresh2017clinical, ma2018general, jin2018treatment, lee2018diagnosis, duan2019clinical, darabi2020taper, choi2020learning, zhang2020time, mcdermott2021comprehensive, sprint2024cogprog} or use multiple modalities with straightforward approaches \cite{xu2018raim, feng2019dcmn, huang2019clinical, khadanga2019using, wang2020utilizing, zhang2020bert, bardak2021improving, yang2021leverage, cui2022automed}, often failing to model the complex, complementary relationships across multimodal data. In contrast, our approach introduces a unified representation learning framework that more effectively bridges semantic and temporal modalities for chronic disease prediction.

\subsection{Longitudinal-based Methods for Disease Prediction}

The temporal characteristics of EHR data present a unique set of challenges. RNN-based models \cite{choi2016retain, choi2016doctor, kwon2018retainvis, ma2017dipole, baytas2017patient, bai2018interpretable, gupta2022obesity} were initially dominant because of their sequential processing capabilities. Models such as RETAIN \cite{choi2016retain}, RetainEX \cite{kwon2018retainvis}, Dipole \cite{ma2017dipole}, and T-LSTM \cite{baytas2017patient} incorporate attention and decay mechanisms to better capture long-term dependencies and irregular visit patterns.

However, these models often struggle with representing complex, irregular time contexts. More recently, transformer-based architectures have emerged as powerful alternatives \cite{li2020behrt, luo2020hitanet, rasmy2021med, li2022hi, yang2023transformehr, moore2024healthrecordbert}. For example, BEHRT \cite{li2020behrt} adapts BERT for EHR sequences via ICD codes, and HiTANet \cite{luo2020hitanet} introduces time-aware transformer modules to model both short- and long-term trends. Med-BERT \cite{rasmy2021med} pretrains transformers on structured diagnosis codes, improving their generalizability to downstream tasks.
Despite these advancements, most transformer models are limited to structured data and do not incorporate unstructured clinical notes or textual lab results. This limits their applicability in real-world clinical settings. Recent works by Cai et al. \cite{cai2024contrastive}, Chung et al. \cite{chung2024large}, and Wang et al. \cite{wang2024recent} emphasize the importance of combining temporal reasoning with semantic understanding through deep multimodal architectures.

Existing models either fail to capture the non-linear progression of chronic conditions or lack the cross-modal integration necessary for comprehensive prediction.
Our proposed cross-modal LLM-Transformer architecture addresses these limitations by fusing clinical notes with temporal embeddings to enable accurate multilabel disease prediction under irregular real-world conditions.

\section{Problem Formulation}
An EHR dataset comprises sequences of patient visits to healthcare facilities. Each visit includes multiple data types such as diagnoses, lab tests, and clinical notes.

\textbf{Visit:} A single visit includes one or more diagnoses. The $t$-th visit is represented as a binary vector $\mathbf{x}_t \in \{0, 1\}^d$, where $d$ is the total number of distinct diseases in the dataset. $x_t^i = 1$ indicates that the $i$-th disease was diagnosed during the $t$-th visit.

\textbf{Visit Sequence:} For patient $p$, the sequence of visits is denoted as $\mathbf{x}_p = (\mathbf{x}_1, \mathbf{x}_2, \dots, \mathbf{x}_T) \in \{0, 1\}^{T \times d}$, where $T$ is the total number of visits.

\textbf{Time Data:} Each visit $\mathbf{x}_t$ is associated with two timestamps, $a_t$ \& $d_t$, indicating the admission and discharge times of the visit, respectively.
\begin{itemize}
    \item \textbf{Time duration} of the $t$-th visit is defined as $\delta^{\text{visit}}_t = d_{t} - a_{t}$.
    \item \textbf{Time interval} between the $t$-th visit and the previous visit is defined as $\delta^{\text{gap}}_t = a_{t} - d_{t-1}$, where $d_{t-1}$ is the discharge time of the $(t-1)$-th visit in the sequence. 
\end{itemize}

\textbf{EHR dataset:} Denoted as $D = \{\mathbf{x}_p \mid p \in P\}$, where $P$ is the set of all patients.

\textbf{Problem formulation:} Given an EHR dataset $D$, the objective is to train a model to predict diseases. For each patient’s visit sequence $\mathbf{x}_u$, the model aims to predict a multilabel vector $\mathbf{y}_p$, where $y_p^i = 1$ indicates the presence of disease $i$.

In addition to multilabel prediction, another often occurring disease prediction method is heart failure prediction. For example, the diagnosis $y^{(T+1)} \in \{0,1\}^d$ represents the ground truth of the diagnostic prediction for the visit $T+1$.

\section{Method}
\subsection{Overview}

\begin{figure}[h!]
\centering
\includegraphics[width=1\textwidth]{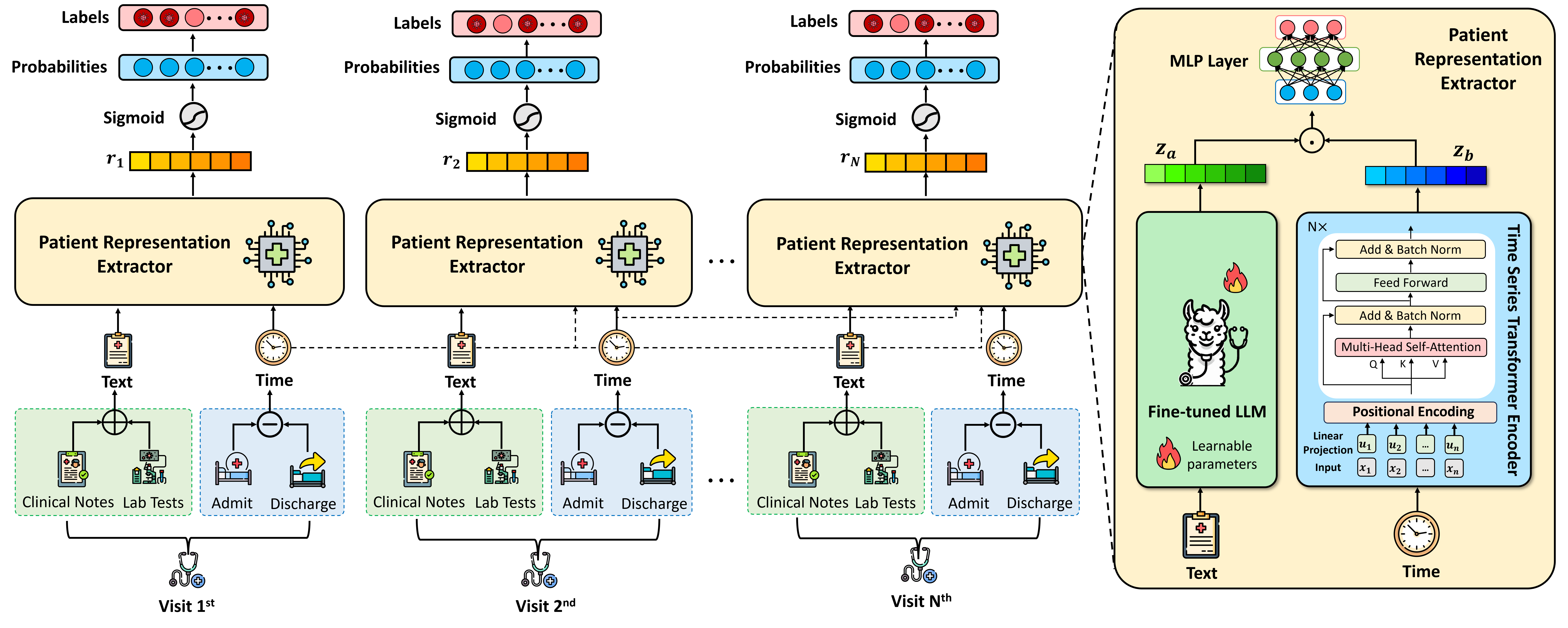}

\caption{The proposed model architecture. The left side depicts the workflow from clinical modality extraction, such as text and time sequences, to patient representation for illness prediction. The right section describes how the Patient Representation Extractor converts laboratory test data to text and merges it with clinical notes that have been run through a fine-tuned LLM to generate clinical embeddings $z_a$. The time series data are also processed via the Time Series Transformer Encoder (TST) to obtain event-sequence embeddings $z_b$. The outputs of these models are then combined and fed into a multi-layer perceptron (MLP) to provide a comprehensive patient representation vector $r$ for multi-disease prediction.}
\label{fig:model}
\end{figure}

Figure \ref{fig:model} illustrates the CURENet architecture, which integrates multiple data sources for clinical predictive modeling. The pipeline starts with the Patient Representation Extraction phase, where different types of patient data, including clinical notes, laboratory results, vital signs, and time series data (admission and discharge times), are processed via specialized extractors to generate structured representations. These representations are then fed into a predictive model for multiple disease prediction tasks. Clinical notes and lab test data are addressed via a fine-tuned LLM, which produces semantic embeddings. Simultaneously, time-stamped data are processed by a Time Series Transformer (TST) encoder to learn temporal event embeddings. Finally, the outputs of these modules are then fused using a multilayer perceptron (MLP) to produce a holistic patient representation vector, enabling robust predictions across multiple chronic conditions.

\subsection{Patient Representation Extractor}
\subsubsection{Fine-tuning the LLMs Module} 
The dataset's textual component consists of clinical notes, which include a variety of fields obtained from physician diagnoses. We utilized the text from the Chief Complaint, Current Illness, Medical History, and Admission Medication sections. These segments are crucial for predicting a patient's condition. To associate these unstructured clinical notes with tabular lab data, we converted the latter into text using a procedure known as tabular feature extraction \cite{thao2024medfuse}. With this technique, abnormal lab results are converted into templated sentences such as ``These are abnormal results recorded: ITEMID$<$ITEMID$>$: $<$VALUE$>$$<$VALUEUOM$>$;
ITEMID$<$ITEMID$>$:$<$VALUE$>$$<$VALUEUOM$>$;...;''. Here, $<$ITEMID$>$ refers to the specific name of the lab test, $<$VALUE$>$ indicates the test values, and $<$VALUEUOM$>$ denotes the units of measure for the test values.

LLMs such as GPT \cite{brown2020language} and LLaMA \cite{touvron2023llama} have recently outperformed traditional deep learning models by leveraging self-supervised learning on massive unlabeled datasets.
These models, often comprising billions of parameters, are adaptable to various tasks and perform well in language processing. Recent investigations have shown the effectiveness of LLMs for diagnostic tasks \cite{zhou2024large}. Our method employs fine-tuning and test-time embedding alignment, which is far more efficient than the pretrained multimodal models proposed by Cai et al. \cite{cai2024contrastive}, which require a substantial amount of time and resources. Motivated by the recent success of fine-tuning language models for classification \cite{devlin2018bert,liu2019roberta}, we refined several LLMs for chronic disease prediction. The Medical-Llama3-8B model \cite{medicalllama3modelcard}, publicly available and refined from Meta-Llama-3-8B \cite{llama3modelcard}, serves as our best-performing backbone. Optimized for clinical reasoning tasks, this model has been fine-tuned on a large-scale medical chatbot corpus, making it well-suited for interpreting domain-specific narratives such as clinical notes and lab-derived textual summaries. \hl{Given an input sequence of clinical notes and formatted lab text $x_{text}$, the LLM generates this information into a high-dimensional hidden representation $h_a$ via its autoregressive transformer decoder:}

\begin{equation}
    h_a = LLM(x_{text})
\end{equation}

\hl{This final hidden vector $h_a\in\mathcal{R}^{d_{\text{LLM}}}$ from the LLM decoder captures rich semantic and contextual information from medical notes, where $d_{\text{LLM}}$ denotes the embedding dimension of the language model. However, this representation has an internal embedding space of the LLM and is not directly aligned with the downstream classification task. To bridge this gap, we apply a linear transformation using a feed-forward neural network to project the LLM embeddings into the disease label space. The resulting output vector $z_a\in\mathcal{R}^d$ contains logits, where each dimension corresponds to the model's score for a specific disease:}
\begin{equation}
    z_a = W_{\text{LLM}}h_a + b_{\text{LLM}}
\end{equation}

\hl{Here, $W_{\text{LLM}}\in\mathcal{R}^{d\times d_{\text{LLM}}}$ and $b_{\text{LLM}}\in\mathcal{R}^d$ denote the learnable weight matrix and bias vector of the classification layer, respectively. This projection aligns the semantic features with the multilabel disease prediction task, converting language-based information into a discriminative space for clinical diagnosis.}

\subsubsection{Time-Series Transformer Module}

\hl{In this study, we specifically focused on admission ($a_t$) and discharge ($d_t$) timestamps as primary temporal signals for modeling patient health trajectories. Rather than incorporating fine-grained physiological time series data, we simplify the timeline into discrete visit events. For each visit, we compute (1) visit duration, defined as $\delta_{\text{visit}} = d_t - a_t$, and (2) inter-visit gap, defined as $\delta_{\text{gap}} = a_t - d_{t-1}$. These scalar values are embedded and concatenated with other visit-level representations to inject temporal awareness into the model.}

\textcolor{black}{While our design omits intra-visit dynamics such as vital or lab trends, these temporal features capture important aspects of chronic disease progression. In particular, longer hospitalization durations may reflect more severe or complex clinical episodes, whereas shorter intervals between visits can indicate disease instability, complications, or poorly controlled chronic conditions. This approach is particularly pertinent given the high prevalence of chronic diseases such as hypertension, diabetes, and congestive heart failure in both datasets (see \ref{fig:diseases}). These conditions are often associated with recurrent hospitalizations and heterogeneous care demands over time.}

\hl{Disease prediction fundamentally relies on modeling temporal progression patterns across patient histories. However, this task is challenging because of the irregular timing of visits and the variability in individual health trajectories. The temporal nature of disease development differs markedly across patients, depending on the complexity of their conditions and their response to treatment. Clinicians typically consider not only a patient's current health status but also the duration and frequency of previous episodes when evaluating risk and planning care. Thus, capturing this longitudinal structure is essential for accurately predicting disease onset and progression.}

To address this, we applied a transformer encoder with temporal masking and padding to process variable-length visit sequences while preserving order. This setup allows the model to focus attention on meaningful time steps, even when visits are irregularly spaced. Although we rely on coarse-grained temporal inputs, this strategy supports a simplified yet effective mechanism for learning progression-aware embeddings, allowing the model to reason over both recent and long-term visit dynamics in a unified representation.
Recurrent Neural Networks (RNNs) and Long Short-Term Memory networks (LSTMs) have been widely employed in sequence modeling tasks, including time series forecasting and language modeling, owing to their ability to process sequential data. However, their architectural constraints make them less effective for specific applications, particularly in disease prediction from clinical time series data. When dealing with an extended time sequence, LSTMs usually fail to depend on the information described at earlier timestamps when later timestamps are processed. This restriction to model forgetting some of the primary essential details of the sequence leads to diminished capabilities in capturing long-term dependencies correctly. 
Transformer architectures have shown remarkable efficacy in a number of fields, including time series analysis and natural language processing, in recent deep learning achievements.
Inspired by \cite{zerveas2021transformer}, we use an encoder-only transformer to extract representations from time sequences, which is ideal for classification problems such as disease prediction, where the task is to infer outcomes from existing data rather than generate new sequences.

The input is modeled as a multivariate time series. Let denote $X=\{x_1,x_2,...,x_w\}$, where $x_t\in \mathcal{R}^m$ contains $m$ features across $w$ time steps. 
The time data are recorded at an hourly level and span a broad range. To standardize it and enhance model generalizability, we applied min-max scaling to obtain the normalized value $\tilde{x_t} \in \mathcal{R}^m$. Each scaled input is projected into a higher-dimensional space $\mathcal{R}^d$:
\begin{equation}
     u_t = W_p \tilde{x}_t + b_p
\end{equation}

where $W_p\in\mathcal{R}^{d\times m}$ is a learnable weight matrix and $b_p\in\mathcal{R}^d$ is the bias term.

To encode the sequential nature of the data, we added positional encoding $p_t\in\mathcal{R}^d$ for the
time step $t$ to the input embeddings:
\begin{equation}
      u_{t}^{'} = u_t + p_t
\end{equation}


This enriched sequence is then processed by the Time Series Transformer (TST) encoder using the self-attention mechanism to capture temporal dependencies:
\begin{equation}
      Attention(Q,K,V) = softmax(\frac{QK^T}{\sqrt{d_k}})V
\end{equation}
where $Q, K, V$ are the query, key, and value matrices derived from the input embeddings $u_{t}^{'} \in\mathcal{R}^d$; $\frac{1}{\sqrt{d_k}}$ is a scaling factor used to stabilize gradients, and $d_k$ is the dimensionality of the key vectors.

To further process the temporal embeddings from TST, we apply multi-head self-attention, allowing the model to focus on both short- and long-term dependencies across visits. After the self-attention layer, each output $x$ is passed through a feed-forward network ($FFN(\cdot)$):

\begin{equation}
    z_t = FFN(x)=ReLU(xW_1 + b_1)W_2 + b_2
\end{equation}
where $W_1\in\mathcal{R}^{d\times d_{hidden}}, W_2\in\mathcal{R}^{d_{hidden}\times d}$ are learnable weight matrices, and biases $b_1 \in\mathcal{R}^{d_{hidden}}, b_2\in\mathcal{R}^{d}$.

In the original transformer architecture proposed by \cite{vaswani2017attention}, layer normalization is applied after both the self-attention mechanism and the feedforward network within each encoder block, leading to substantial performance improvements over batch normalization. In our work, however, we adopt batch normalization instead. This decision is motivated by its superior performance in handling outliers in time series data, which are more prominent in clinical settings than in NLP. Furthermore, the relatively poor performance of batch normalization in NLP has largely been linked to the wide variability in sequence lengths \cite{shen2020powernorm}, a problem that is significantly reduced in our datasets.

Finally, a padding mask is applied to ensure that the model disregards padded elements during the attention computation, allowing it to handle sequences of varying lengths effectively. This results in the modified sequence representation denoted as $\tilde{z}\in\mathcal{R}^{d}$. The final encoded temporal representation $z_b\in\mathcal{R}^{d}$ is obtained by aggregating the sequence embeddings produced by the encoder.

\begin{equation}
     z_b = W_{o}\tilde{z} + b_o
\end{equation}
where $W_o\in\mathcal{R}^{d\times d}$ is a learnable weight matrix and $b_o\in\mathcal{R}^d$ is the bias.

\subsubsection{Holistic Patient Representation}
Unlike prior hierarchical transformer models (e.g., HiTANet \cite{luo2020hitanet}), our architecture supports fine-grained fusion across semantic and temporal dimensions by aligning LLM-derived embeddings with time-aware transformer outputs. In this framework, CURENet generates a unified representation through a two-stream patient representation extractor. We employed cross-modal representation learning, wherein the fine-tuned medical LLM component extracts semantic features $z_a$ from unstructured clinical notes and text-converted lab results, leveraging extensive medical knowledge encoded during pre-training. Simultaneously, the Time Series Transformer Encoder (TST) generates temporal embeddings $z_b$ that encapsulate the longitudinal dynamics of physiological parameters, such as visit duration and inter-visit intervals.

\hl{To integrate these heterogeneous information sources, we concatenate the two embeddings (semantic ($z_a\in\mathcal{R}^d$) and temporal ($z_b\in\mathcal{R}^d$)) into a single joint representation vector $z_{in}\in\mathcal{R}^{d+d}$. This operation is denoted by $||$, which represents vector concatenation:}

\begin{equation}\label{eq:concat}
z_{in} = [z_a || z_b]
\end{equation}

\hl{Here, ``$||$" means that the values of $z_a$ and $z_b$ are stacked along the feature dimension, preserving all information from both modalities in a single unified vector. This enables the downstream model to jointly reason over both semantic content (from clinical notes and lab text) and temporal patterns (from visit timelines).}
 
\hl{The resulting fused vector $z_{in}$ is then passed through a multilayer perceptron (MLP), which enables the model to learn nonlinear interactions between these modalities and project the combined information into a space optimized $h\in\mathcal{R}^d$ for disease prediction. The MLP applies a series of transformations, beginning with the following:}
\begin{equation}
h = ReLU(W_1 z_{in} + b_1)
\end{equation}
where $W_1\in\mathcal{R}^{d\times (d+d)}$ is a learnable weight matrix and $b_1\in\mathcal{R}^d$ is the bias.

Further hidden layers can be incorporated, with the final layer producing the comprehensive patient representation $z\in\mathcal{R}^d$:
\begin{equation}
z = W_{out}h + b_{out}
\end{equation}
where $W_{out}\in\mathcal{R}^{d\times d}$ is a learnable weight matrix and $b_{out}\in\mathcal{R}^d$ is the bias.

\hl{This joint representation is optimized end-to-end, enabling the model to learn both the semantic richness from textual sources and the longitudinal structure from temporal sequences. This multimodal fusion captures patterns that are not easily detected by unimodal methods. Furthermore, the architecture implements hierarchical temporal modeling to address the inherent multiscale characteristics of disease progression.}





\subsection{Prediction}
After obtaining the holistic patient representation $z$, the output $\hat{O}=\sigma(z)$ can be produced to derive classification probabilities with scaling by a sigmoid function $\sigma$. The 10-chronic-disease prediction combination was then derived by dichotomizing the model output $\hat{O}$ to 0 or 1 with a threshold $\delta=0.5$.

\subsection{Model Training}
In clinical multilabel classification tasks, label distributions are typically imbalanced and interdependent, making model optimization challenging. Relying on a single loss function may not sufficiently capture the complexity of the prediction space. The commonly used Binary Cross-Entropy (BCE) loss function captures individual label probabilities but treats each independently, often overlooking the relative ranking between positive and negative classes. In contrast, multilabel hinge loss introduces a margin-based constraint that penalizes the model when negative labels are scored higher than positive labels are, thereby improving label discrimination and ranking performance. To explore the advantages of these objectives, we adopt a hybrid loss function that linearly combines BCE and hinge loss. This composite approach balances robust label prediction with better ranking performance \cite{nam2014large, feng2019dcmn, cai2024contrastive}.


\textbf{Binary Cross-Entropy Loss} was employed to evaluate our model's classification performance. This loss quantified the distinction between the actual value and the expected result. The BCE formula is -

\begin{equation}
    \mathcal{L}_{bce} = - \sum_{p=1}^{|\mathcal{P}|} \left[ o_p^\top \log(\hat{o}_p) + (1 - o_p)^\top \log(1 - \hat{o}_p) \right]
\end{equation}

\textbf{Multilabel Hinge Loss} was applied to determine the accuracy of our multilabel classification model. This loss function assesses the disparity between the predicted and actual labels. The formula for this loss is as follows:

\begin{equation}
    \mathcal{L}_{multi}=\sum_{i,j:o_{i}=1, o_{j}=0}\frac{max(0,1-(\hat{o}_{i}-\hat{o}_{j}))}{\left | \mathcal{P} \right |}
\end{equation}

where \( |\mathcal{P}| \) represents the total number of patient samples, \( o_p \) is the ground truth label for the \( p \)-th patient, and \( \hat{o}_p \) denotes the predicted probability for the same patient. 

\textbf{Overall Loss Function.} One popular technique for training with multiple loss functions is to compute the weighted sum of the terms that measure the loss \cite{dosovitskiy2019you}. The two previously specified loss functions were linearly combined to form the overall loss function:
\begin{equation}
   \mathcal{L}(\theta)=(\alpha \mathcal{L}_{bce}+(1-\alpha)\mathcal{L}_{multi})
\end{equation}
where $\alpha$ is a hyperparameter. 


\hl{To balance the trade-off between probabilistic calibration and label ranking, we combined Binary Cross-Entropy loss ($\mathcal{L}_{bce}$) with multilabel hinge loss ($\mathcal{L}_{multi}$) via a weighted coefficient $\alpha \in [0,1]$. We set $\alpha = 0.95$ on the basic of a grid search over the validation set, prioritizing $\mathcal{L}_{bce}$ to ensure robust per-label classification under label imbalance, while still benefiting from the margin-based discrimination provided by $\mathcal{L}_{multi}$. This formulation is particularly suited for clinical multilabel prediction, where accurate probability estimation and effective ranking are both essential, especially for underrepresented but critical conditions.}


\hl{To further improve model stability and generalizability, we applied min-max scaling to temporal features (visit duration and inter-visit gap), and used padding with temporal masking to handle variable-length visit sequences without introducing spurious attention. We also filtered the dataset to include only patients with at least two visits and ensured full de-identification of all EHR data.
These combined strategies, across loss design and data preprocessing, enhance model robustness to class imbalance, temporal irregularity, and noise. The full training pipeline is summarized in Algorithm~\ref{algo:curenet}.}


\section{EXPERIMENTS}
In this part, we evaluate CURENet by answering five key research questions:
\begin{itemize}
    \item \textbf{RQ1:} How does CURENet compare with existing, cutting-edge general and medical large language models for diagnostic prediction?
    \item \textbf{RQ2:} How effective is CURENet in heart failure prediction (binary classification)?
    \item \textbf{RQ3:} What are the effects of multimodal input data on model performance?
    \item \textbf{RQ4:} How can a comprehensive case study help evaluate and explain the diagnostic results?
    \item \textbf{RQ5:} How do CURENet's learned disease embeddings compare with those of baselines?
\end{itemize}

\RestyleAlgo{ruled} 
\SetKwInput{KwInput}{Input} 
\SetKwComment{Comment}{/* }{ */} 
\begin{algorithm}[ht!]
\caption{Training process of \textbf{CURENet}}\label{algo:curenet}
\KwInput{Training set $X^{\text{train}}$;}
Constructing a Time Series Transformer (TST) Encoder Module\;
Initialize parameters $E_{d}$ for time duration\;
Load pretrained weights $W_{t}$ to the LLMs module\;

\For{$epoch \gets 1$ \KwTo $n\_\text{epoch}$}{
    \For{$patient \gets 1$ \KwTo $\left|X^{\text{train}}\right|$}{
        Get patient $p$’s history, $X_{p}$\;
        
        \For{$visit \gets 1$ \KwTo $\left|X_{p}\right|$}{
            Select the time duration sequence for visits $1$ to $t$, $x_{p}^{(:t)}$\;
            Select clinical notes at the $t$-th visit of patient $p$, $text_{p}^{(t)}$\;
            Select lab tests at the $t$-th visit of patient $p$, $labtest_{p}^{(t)}$\;
            Label of the $t$-th visit of patient $p$, $o_{p}^{(t)}$\;
            $z_a \gets \text{\textit{TSTEncoder}}\left(x_{p}^{(:t)}\right)$\;
            $z_b \gets \text{\textit{LLMs}}\left([text_{p}^{(t)}, labtest_{p}^{(t)}]\right)$\;
            $r \gets \text{\textit{Fusion}}\left(z_a, z_b\right)$\;
            Generate disease prediction $\hat{o}_{p}^{(t)}$\;
            Compute $\mathcal{L}_{\text{bce}}$ and $\mathcal{L}_{\text{multi}}$ \Comment*[r]{Losses}
            
            $\mathcal{L} \gets \alpha \mathcal{L}_{\text{bce}} + (1 - \alpha) \mathcal{L}_{\text{multi}}$\;
            Optimize parameters on the basic of $\mathcal{L}$\;
        }
    }
}
\end{algorithm}

\subsection{Experimental Setup}

\subsubsection{Tasks}
We performed two tasks: diagnostic prediction via multilabel classification and heart failure prediction via binary classification. In disease prediction, the model forecasts possible diseases for visit $T+1$ on the basic of previous $T$ visits. Heart failure prediction is the forecast that a patient will be diagnosed with heart failure at visit $T+1$ based on previous $T$ visits.

\renewcommand{\arraystretch}{1.25}%
\begin{table}[h]
\centering
\caption{The number of samples in the datasets.}
\label{tab:dt}
\begin{tabular}{lcc}
\toprule
\textbf{} & \textbf{MIMIC-III Dataset} & \textbf{FEMH Dataset} \\
\hline
\textbf{\#Training Samples} & 7,629 &  10,676\\
\textbf{\#Testing Samples} & 1,928 & 2,670 \\
\bottomrule
\end{tabular}
\end{table}

\subsubsection{Datasets}
We utilized two real-world EHR datasets, MIMIC-III \cite{mimic3} and FEMH, to assess the performance of the approaches under comparison. From 2017 to 2021, we obtained five years of EHRs from the Far Eastern Memorial Hospital (FEMH) in Taiwan. There are over 1,505 lab test items, 387,392 lab results, and 1,420,596 clinical notes in the collection. The study was authorized by the FEMH Research Ethics Review Committee\footnote{https://www.femh-irb.org/}, and all the data were de-identified. We selected patients with at least two documented visits for the MIMIC-III and FEMH datasets. Table \ref{tab:dt} presents the training and testing sample distributions with the total number of samples in each dataset.
\hl{For both the MIMIC-III and FEMH datasets, we adopted a patient-level split to ensure that visits from the same individual did not appear in both the training and testing sets, thereby avoiding data leakage. The datasets were randomly partitioned into training (80\%) and testing (20\%) subsets. We chose this scheme to simulate a realistic deployment setting in which a model must operate on entirely new patients.}

\begin{figure}[ht!]
\centering
\includegraphics[width=\textwidth]{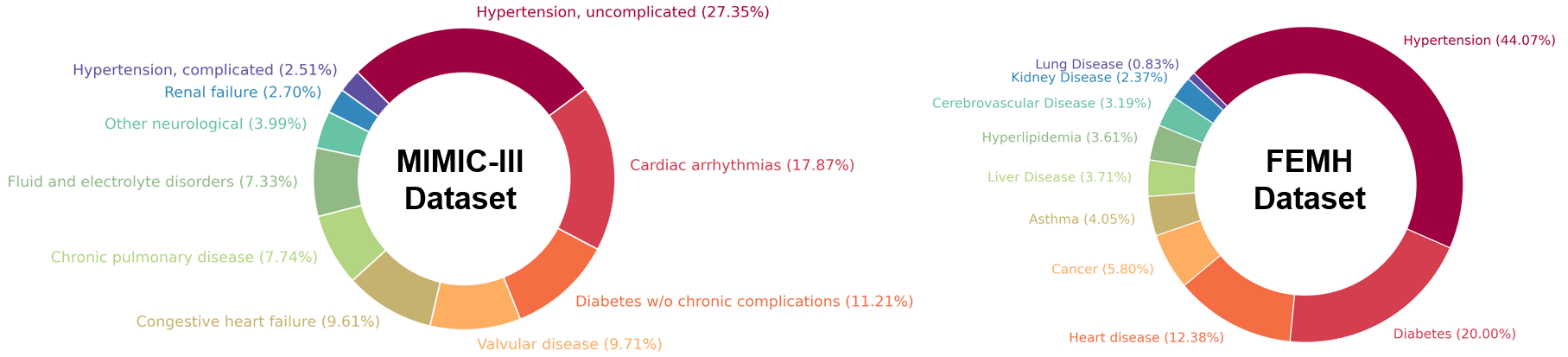}
\caption{Top 10 disease statistics in MIMIC-III and FEMH datasets.}
\label{fig:diseases}
\end{figure}

For the MIMIC-III dataset, we selected the ten most prevalent conditions for the multilabel classification task: Hypertension, uncomplicated (HTN), Cardiac arrhythmias (ARRHY), Diabetes w/o chronic complications (DM), Valvular disease (VALVE), Congestive heart failure (CHL), Chronic pulmonary disease (CHRNLUNG), Fluid and electrolyte disorders (LYTES), Other neurological (NEURO), Renal failure (RENLFAIL), and Hypertension, complicated (HTNCX). These categories encompass a broad range of cardiovascular, metabolic, pulmonary, renal, and neurological conditions, reflecting the predominant causes of long-term morbidity in intensive care cohorts.

For the FEMH dataset, the ten most common conditions were identified as follows: Hypertension (HTN), Diabetes (DM), Heart disease (CHF), Cancer (CANCER), Asthma (ASTHMA), Liver Disease (LIVER), Hyperlipidemia (HYPERLIP), Cerebrovascular (CEREBVSC), Kidney Disease (KIDNEY), and Lung Disease (LUNG). \hl{This distribution is dominated by hypertension and diabetes, with additional representations of cardiovascular, oncological, respiratory, and metabolic disorders, which is consistent with national epidemiological trends in Taiwan}. Figure \ref{fig:diseases} provides a summary of the top 10 disease statistics for both the MIMIC-III and FEMH datasets.

\hl{While these sets reflect different healthcare environments, both exhibit a concentration of chronic conditions that drive long-term patient management needs.
Selecting these high-prevalence conditions ensures adequate sample sizes for each label while maintaining clinical relevance.
This targeted selection also enables meaningful multilabel prediction, where co-occurrence patterns among frequent diseases can be leveraged to improve predictive accuracy.}

\subsubsection{Metrics}
We evaluated all the models using established multilabel classification metrics, including precision, recall, accuracy, weighted-average, and macro-average F1-scores \cite{hossin2015review, palacio2019evaluation}, to capture diverse performance aspects.
\hl{Precision, recall, and F1-scores address the trade-off between false positives and false negatives, a critical concern in clinical contexts. Weighted-average F1 emphasizes common diagnoses, whereas macro-average F1 ensures fair assessment of rare conditions. Accuracy, although less robust in imbalanced data, was included for interpretability and comparability with prior work.
Additionally, we employed the ranking-based metrics Recall@k and NDCG@k \cite{tan2023enhancing, lu2022context} to evaluate the quality of the top-k predictions, aligning with clinical practice where physicians review a shortlist of likely diagnoses.
Recall@k measures the presence of relevant diagnoses in the top-k outputs, whereas NDCG@k further considers their ranking positions.}

\begin{itemize}
    \item \textbf{Recall@k} quantifies the average proportion of true relevant illnesses caught by the top $k$ predictions for each patient throughout the test set.
    \begin{equation}
        \text{Recall@k} = \frac{1}{\left | \mathcal{P} \right |} \sum_{i=1}^{\left | \mathcal{P} \right |} \frac{|D_i \cap \hat{D}_i^{(k)}|}{|D_i|}
    \end{equation}
    where $\left | \mathcal{P} \right |$ represents the total number of patients in the test set, $D_i$ is the set of true, relevant diseases (actual diagnoses) for the $i$-th patient, and $\hat{D}_i^{(k)}$ denotes the set of the top $k$ predicted diseases for the $i$-th patient, based on the model's ranking.
    
    \item \textbf{NDCG@k} is an assessment metric in ranking-based recommendation systems. The greater weight given to the diseases ranked higher is used to evaluate the quality of the top $k$ predictions, considering the relevance of the suggested diseases and their placements within the ranking. The Discounted Cumulative Gain (DCG) is then computed by adding these relevant ratings and discounting each value based on the disease's position in the prioritized list. To obtain the final NDCG score, the DCG is normalized by dividing it by the Ideal DCG (IDCG), which reflects the greatest DCG achievable for the provided list.
    \begin{equation}
        \text{DCG@k} = \sum_{i=1}^{k} \frac{2^{\text{rel}_i} - 1}{\log_2(i + 1)}
    \end{equation}
    \begin{equation}
        \text{IDCG@k} = \sum_{i=1}^{|\text{REL}|} \frac{2^{\text{rel}_i} - 1}{\log_2(i + 1)}
    \end{equation}
    \begin{equation}
        \text{NDCG@k} = \frac{\text{DCG@k}}{\text{IDCG@k}}
    \end{equation}
    where $k$ denotes the number of anticipated diagnoses, and $i$ denotes a query index in the test set. The relevance level is indicated by $\text{rel}_i$, where 1 indicates relevance and 0 indicates irrelevance. The collection of the top $k$ diagnoses, arranged in descending order of relevance, is called $|\text{REL}|$.
\end{itemize}

\subsection{Implementation Details}
This section outlines the experimental setup, including data processing, model configurations, and parameter settings, along with details on the sampling strategy used in the testing phase. The experiments were conducted on a Linux workstation with 256GB RAM, 112 CPU cores, and a 48GB NVIDIA RTX 6000 Ada Generation GPU.
\hl{Despite the high-end setup, the use of 4-bit NF4 quantization~\cite{dettmers2023qlora} combined with LoRA adapters~\cite{hu2022lora} significantly reduces memory usage, enabling inference on 8-12 GB GPUs and fine-tuning on 16-24 GB GPUs with gradient checkpointing and a small batch size of 8.}
The dataset was randomly split into training and validation sets in a $\frac{4}{5}:\frac{1}{5}$ ratio. Our implementation leverages a Time-Series Transformer (TST) module, with a fusion setup combining outputs from both an LLM and the TST module by concatenating their embeddings. The model was trained with a learning rate of 1e-4, focusing on two main features: time duration and time interval, to classify the data into 10 distinct chronic diseases. In the TST module, the model handles sequences with a maximum length of 16, using embeddings of 64 dimensions. The attention mechanism employs 8 heads across 3 transformer layers, with a feedforward network size of 256. To enhance model stability, we applied 10\% dropout regularization and fixed positional encoding. The activation function is the GELU, which provides smooth and robust gradients, and Layer Normalization stabilizes learning. Notably, all the parameters are unfrozen, allowing them to be updated during training for dynamic learning. The training process of CURENet was finished in 10 epochs.

\subsection{Experimental Results}
\subsubsection{Overall Performance Comparison (RQ1)}
Table \ref{tab:results} illustrates that CURENet outperforms the baseline models on the FEMH and MIMIC-III datasets. Our approach performs better than baseline models do on all important criteria, including BERT \cite{devlin2018bert}, Mistral-7B-v0.1 \cite{jiang2023mistral}, Llama-2-7B-hf \cite{touvron2023llama}, Meta-Llama2-13B \cite{llama2modelcard}, Meta-Llama3-8B \cite{llama3modelcard}, and Medical-Llama3-8B \cite{medicalllama3modelcard}. These findings demonstrate that our suggested multi-disease personalized approach may accurately predict diagnoses, addressing RQ1. For the test set, our model performs 1.49\% better in terms of the macro F1 score, and similar trends are observed for the other metrics. For the FEMH dataset, CURENet outperforms the rival LoRA Medical-Llama3-8B model but with more modest and consistent gains across measures. CURENet's strong performance improved by approximately 2\% to 4\% on both datasets. Notably, medical LLMs consistently outperform general-purpose LLMs across various evaluation metrics such as accuracy, precision, recall, and the F1 macro. The advantages of models designed especially for medical applications are that they are probably more suited to manage the complex language and subtle information in clinical data. Our findings show CURENet's robustness and reliability in predicting healthcare tasks, even when dealing with clinical data with complex temporal relationships. It is particularly suitable for disease prediction and patient risk assessment. 

\renewcommand{\arraystretch}{1.25}%
\begin{table*}[htbp!]
\caption{Performance comparison of the MIMIC-III and FEMH datasets (training/validation)}
\label{tab:results}
\centering
\resizebox{\textwidth}{!}{%
\begin{tabular}{l|ccccc}
\toprule
\multirow{2}{*}{\textbf{Model}}         & \textbf{Precision} & \textbf{Recall} & \textbf{F1 macro} & \textbf{F1 weighted} & \textbf{Accuracy} \\ \cmidrule{2-6}
                       & \multicolumn{5}{c}{\textbf{MIMIC-III}}                                                                       \\ \hline
BERT                   & 0.8333 / 0.6790    & 0.2000 / 0.2000 & 0.1818 / 0.1618   & 0.3686 / 0.3162      & 0.6515 / 0.2692   \\
LoRA Mistral-7B-v0.1   & 0.8759 / 0.8616    & 0.8459 / 0.8289 & 0.8449 / 0.8274   & 0.9007 / 0.8886      & 0.9089 / 0.8974   \\
LoRA Llama-2-7B-hf     & 0.8828 / 0.8585    & 0.8592 / 0.8364 & 0.8559 / 0.8301   & 0.9097 / 0.8924      & 0.9168 / 0.9004   \\
LoRA Meta-Llama2-13B   & 0.9153 / 0.8732    & 0.8852 / 0.8430 & 0.8874 / 0.8414   & 0.9297 / 0.8990      & 0.9363 / 0.9071   \\
LoRA Meta-Llama3-8B    & 0.8899 / 0.8667    & 0.8569 / 0.8306 & 0.8579 / 0.8305   & 0.9121 / 0.8935      & 0.9211 / 0.9040   \\
LoRA Medical-Llama3-8B & 0.9283 / 0.8807    & 0.9008 / 0.8474 & 0.9026 / 0.8466   & 0.9367 / 0.9003      & 0.9417 / 0.9068   \\
\textbf{CURENet (Ours)} &
  \textbf{0.9370 / 0.8839} &
  \textbf{0.9100 / 0.8571} &
  \textbf{0.9123 / 0.8551} &
  \textbf{0.9450 / 0.9110} &
  \textbf{0.9492 / 0.9166} \\ \hline
                       & \multicolumn{5}{c}{\textbf{FEMH}}                                                                            \\ \hline
LoRA Medical-Llama3-8B &
  \multicolumn{1}{l}{0.8702 / 0.8691} &
  \multicolumn{1}{l}{0.8496 / 0.8478} &
  \multicolumn{1}{l}{0.8453 / 0.8435} &
  \multicolumn{1}{l}{0.9182 / 0.9167} &
  \multicolumn{1}{l}{0.9267 / 0.9252} \\
\textbf{CURENet (Ours)} &
  \multicolumn{1}{l}{\textbf{0.8939 / 0.8862}} &
  \multicolumn{1}{l}{\textbf{0.8872 / 0.8792}} &
  \multicolumn{1}{l}{\textbf{0.8800 / 0.8720}} &
  \multicolumn{1}{l}{\textbf{0.9417 / 0.9377}} &
  \multicolumn{1}{l}{\textbf{0.9458 / 0.9425}} \\ \bottomrule
\end{tabular}
}
\end{table*}

To ensure a fair and unbiased evaluation, we do not directly compare CURENet with models that rely solely on structured medical codes (e.g., ICD-based models such as BEHRT \cite{li2020behrt} and Med-BERT \cite{rasmy2021med}). These models are fundamentally limited to single-modality inputs and are not designed to process the diverse data types (unstructured clinical notes, textual lab results, and temporal visit data) that CURENet integrates. Their restricted input space makes them incompatible with multimodal frameworks such as ours, and a direct comparison would conflate differences in model capacity with differences in input richness. Instead, we focus on demonstrating CURENet’s strength in modeling complex, real-world clinical scenarios that require comprehensive multimodal reasoning.

\begin{figure}[h!]
    \centering
    \includegraphics[width=\textwidth, trim={0 0.4cm 0 0.4cm},clip]{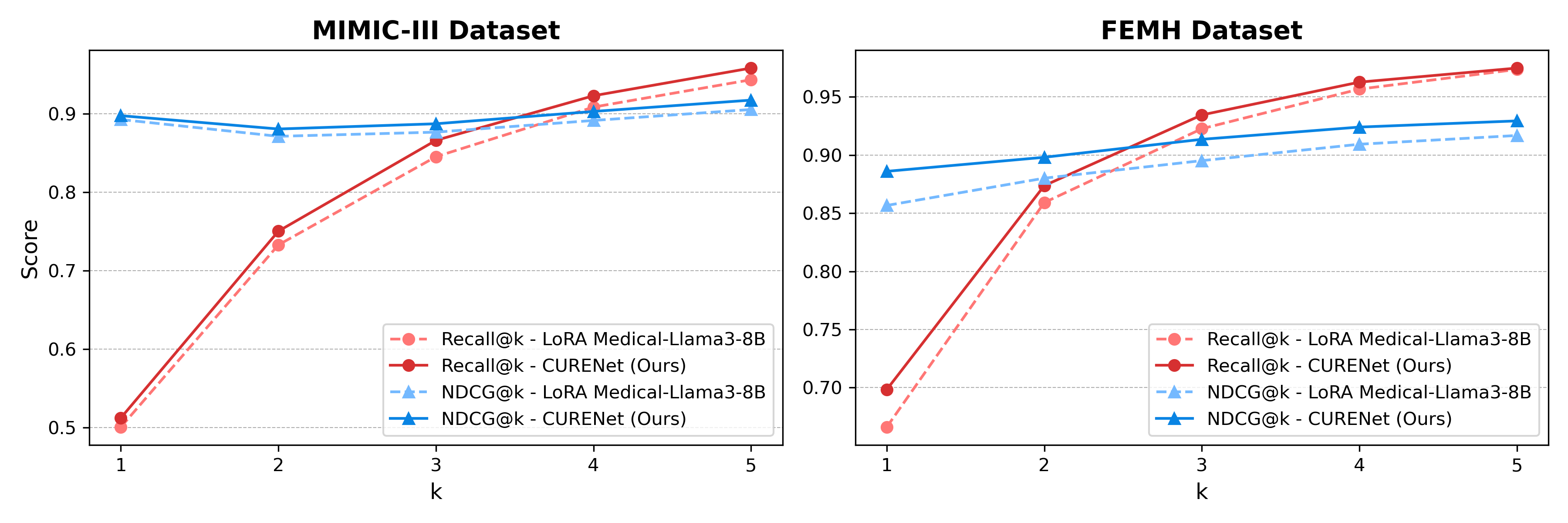}
    \caption{Performance comparison on the MIMIC-III and FEMH datasets using Recall@$k$ and NDCG@$k$}
    \label{fig:metric@k}
\end{figure}
Figure \ref{fig:metric@k} presents two line graphs comparing the performance of the LoRA Medical-Llama3-8B and CURENet (ours) models on the MIMIC-III and FEMH datasets via the Recall@k and NDCG@k metrics, where $k$ ranges from 1 to 5. In the MIMIC-III dataset, the Recall@k results indicate that CURENet outperforms LoRA Medical-Llama3-8B consistently as $k$ increases, reflecting stronger retrieval performance. Moreover, the NDCG@k scores remain relatively high and stable, with LoRA Medical-Llama3-8B slightly outperforming CURENet in terms of ranking performance. In the FEMH dataset, with scores ranging from 0.7 to 0.95, CURENet clearly improves the Recall@k as $k$ increases, outperforming LoRA Medical-Llama3-8B at higher $k$ values. Similarly, the NDCG@3 and NDCG@5 corroborate the effectiveness of our model in ranking retrieved items with significant gains in the normalized discounted cumulative gain. In the FEMH dataset, CURENet remains superior in terms of recall and ranking performance. Overall, the findings suggest that CURENet has stronger recall capabilities on both datasets, whereas LoRA Medical-Llama3-8B maintains a marginal advantage in ranking performance. The robustness and effectiveness of CURENet in enhancing diverse medical datasets are promising approaches for medical information retrieval tasks.

\subsubsection{Heart Failure Prediction (RQ2)}
\begin{figure}[htbp!]
    \centering
    \includegraphics[width=0.9\textwidth, trim={3.6cm 1cm 1.5cm 0.9cm},clip]{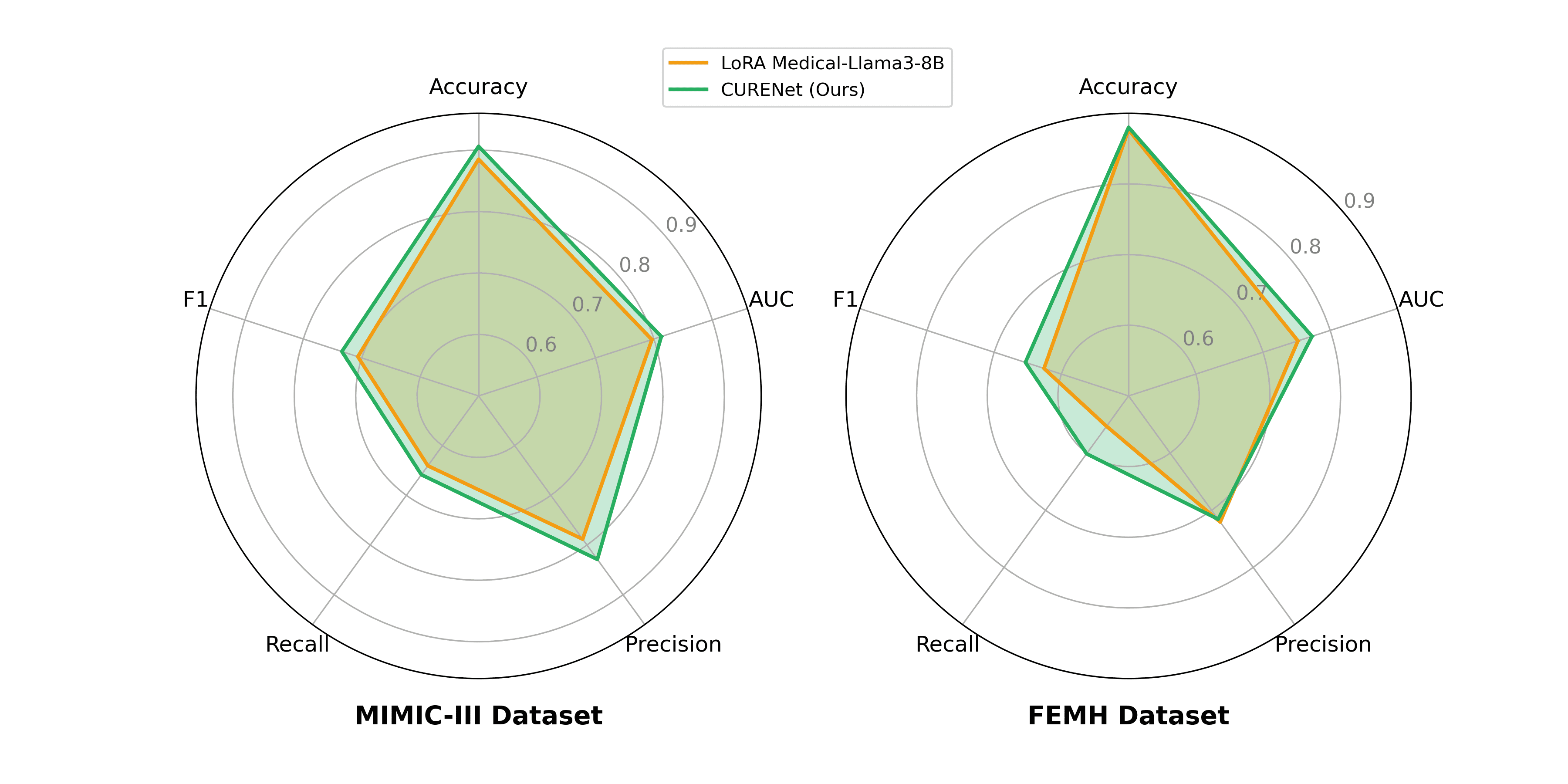}
    \caption{Performance comparison of Heart Failure Prediction on MIMIC-III and FEMH.}
    \label{fig:CHF}
\end{figure}

Figure \ref{fig:CHF} compares the performance of two heart failure prediction models, LoRA Medical-Llama3-8B, and CURENet, on the MIMIC-III and FEMH datasets in terms of the recall, AUC, and F1 score. While both models have strengths, CURENet’s traditional performance across both datasets, especially in terms of recall and F1 score, makes it a more reliable choice for heart failure prediction. Its ability to maintain high sensitivity while balancing precision is critical for minimizing risks in medical decision-making. CURENet is better at balancing sensitivity and precision on the MIMIC-III dataset. On the basic of F1 scores, the model restrains the trade-off between false positives and false negatives more effectively, which is crucial in clinical contexts where incorrect predictions could have serious implications. For the FEMH dataset, CURENet consistently outperforms all the other metrics. Our model is effectively adaptable and generalizable to various datasets, which is suitable for practical applications where data variability is expected. The higher recall achieved by CURENet on FEMH suggests that it is more useful at determining cases of heart failure, reducing the risk of missing critical diagnoses. Additionally, CURENet’s strong ability to distinguish between patients with and without heart failure is an essential aspect in building trust for deployment in clinical settings.

\subsection{Ablation Study (RQ3)}

{\renewcommand{\arraystretch}{1.25}%
\begin{table}[h!]
\centering
\caption{Ablation Study of CURENet for Multidisease Prediction on the MIMIC-III Dataset}
\label{tab:ablation1}
\begin{tabular}{l|ccccc}
\toprule
\multirow{2}{*}{\textbf{Model}} & \multicolumn{5}{c}{\textbf{Classification Metrics}}                                                                            \\ \cmidrule{2-6}
         & \textbf{Precision} & \textbf{Recall} & \textbf{F1 macro} & \textbf{F1 weighted} & \textbf{Accuracy} \\ \hline
w/o TEXT & 0.7144	& 0.6686 & 0.6634 & 0.7903 & 0.8097   \\
w/o LABTEXT\, & 0.8399 & 0.8209 & 0.8124 & 0.8838 & 0.8910   \\
\textbf{CURENet} &
  \textbf{0.8839} &
  \textbf{0.8571} &
  \textbf{0.8551} &
  \textbf{0.9110} &
  \textbf{0.9166} \\ \hline
    & \multicolumn{5}{c}{\textbf{Ranking Metrics}} \\ \cmidrule{2-6}
\textbf{}         & \textbf{Recall@3} & \textbf{NDCG@3} & \textbf{Recall@5} & \textbf{NDCG@5} \\ \hline
w/o TEXT & 0.6417 & 0.6364 & 0.8023 & 0.7058   \\
w/o LABTEXT\, & 0.8190 & 0.8367 & 0.9283 & 0.8759   \\
\textbf{CURENet} &
  \textbf{0.8660} & \textbf{0.8871} & \textbf{0.9580} & \textbf{0.9172} \\ \bottomrule
\end{tabular}
\end{table}

{\renewcommand{\arraystretch}{1.25}%
\begin{table}[h!]
\centering
\caption{Ablation Study of CURENet for Heart Failure Prediction on the MIMIC-III Dataset}
\label{tab:ablation2}
\begin{tabular}{l|ccccc}
\toprule
\multirow{2}{*}{\textbf{Model}} & \multicolumn{5}{c}{\textbf{Classification Metrics}}                                                                            \\ \cmidrule{2-6}
         & \textbf{Precision} & \textbf{Recall} & \textbf{F1 macro} & \textbf{F1 weighted} & \textbf{Accuracy} \\ \hline
w/o TEXT & 0.5556 & 0.0348 & 0.5141 & 0.0662 & 0.7925   \\
w/o LABTEXT\, & 0.7756 & 0.5540 & 0.7574 & 0.6463 & 0.8810   \\
\textbf{CURENet} &
  \textbf{0.8289} & \textbf{0.6585} & \textbf{0.8127} & \textbf{0.7340} & \textbf{0.9063} \\ \bottomrule
\end{tabular}
\end{table}

\hl{We conducted ablation research (as shown in Table \ref{tab:ablation1}-\ref{tab:ablation2})} on the MIMIC-III dataset, which revealed that our full model's performance is much better when clinical notes (TEXT) and lab tests (LABTEXT) are combined. \hl{For the Multi-Disease Prediction task (Table \ref{tab:ablation1})}, CURENet performs better than removing either TEXT or LABTEXT does, obtaining the greatest accuracy (0.8839), recall (0.8571), and F1 score (0.8551). \hl{For the Heart Failure Prediction task (Table \ref{tab:ablation2})}, CURENet outperforms models that use only a single modality. These findings demonstrate how complex patient health issues extend beyond the purview of a single modality and how multimodal data integration is critical for more dependable and accurate predictions of model performance.

\subsection{Case Study (RQ4)}\label{sec:case_study}

\definecolor{green}{rgb}{0.08, 0.69, 0.1}
{\renewcommand{\arraystretch}{1.25}%
\begin{table}[htbp!]
\caption{Predictive diagnoses for patients from the MIMIC-III dataset.}
\label{tab:case-study}
\centering
\begin{tabular}{lll}
\toprule
& \textbf{Medical-Llama3-8B's diagnoses}  & \textbf{CURENet's diagnoses} \\ \hline
\textbf{Visit 1}     & \begin{tabular}[c]{@{}l@{}}ARRHY Cardiac arrhythmias\\ CHF Congestive heart failure \\ \;\end{tabular}                                                             & \begin{tabular}[c]{@{}l@{}}ARRHY Cardiac arrhythmias \\ CHF Congestive heart failure\\ \textcolor{green}{HTN Hypertension, uncomplicated}\end{tabular}                        \\ \hline
\textbf{Visit 2}     & \begin{tabular}[c]{@{}l@{}}\textcolor{red}{HTN Hypertension, uncomplicated}\\ ARRHY Cardiac arrhythmias\\ CHF Congestive heart failure\end{tabular}                                                   & \begin{tabular}[c]{@{}l@{}}HTN Hypertension, uncomplicated\\ ARRY Cardiac arrhythmias \\ CHF Congestive heart failure\end{tabular}                           \\ \hline
\textbf{Visit 3}     & \begin{tabular}[c]{@{}l@{}}\textcolor{red}{HTN Hypertension, uncomplicated}\\ ARRHY Cardiac arrhythmias\\ VALVE Valvular disease\\ CHF Congestive heart failure\end{tabular} & \begin{tabular}[c]{@{}l@{}}HTN Hypertension, uncomplicated\\ ARRHY Cardiac arrhythmias\\ VALVE Valvular disease \\ CHF Congestive heart failure\end{tabular} \\ 
\bottomrule
\end{tabular}
\end{table}

We conducted a case study comparing the predictive performance of the baseline model (LoRA Medical-Llama3-8B) and our proposed model, CURENet, across three patient visits from the MIMIC-III dataset (as shown in Table \ref{tab:case-study}).
The diagnoses predicted by our proposed model are shown, with black representing true positives, red indicating false negatives, and green highlighting false positives, which are diagnoses predicted by the model but not present in the ground-truth diagnosis sets. This illustrates our model's ability to recognize and forecast the progression of chronic conditions over time, even when potential conditions not documented initially as identified.
In addition, the analysis illustrates not only the accuracy of CURENet's predictions but also its potential for interpretability in a clinical context.

Unusual visit patterns are linked to the use of time-ordered data for the purpose of modeling disease progression. Hypertension and congestive heart failure are severe cases that develop slowly and thus deserve attention from historical data, even after some time has elapsed. Furthermore, it adapts to inconsistency in visit frequencies for chronic versus minor conditions, treating the present condition more seriously for acute conditions at frequent visits and maintaining the relevance of the old record at sparse visits. The model distinguishes relevant and irrelevant visits on the basic of mainly chronic conditions, giving proper weight to wider gaps in attendance.

\hl{False positives may reflect the model’s ability to detect underlying conditions that are present but not yet documented in the ground truth, underscoring its clinical sensitivity.
For example, CURENet predicted ``Hypertension, uncomplicated" during Visit 1, even though this diagnosis had not yet appeared in the ground truth. Remarkably, this condition was formally diagnosed at Visit 2, suggesting that the model inferred a latent clinical signal aligned with the eventual physician diagnosis.
This illustrates CURENet’s ability to integrate textual and temporal cues to identify early signals of chronic disease progression, which aligns with longitudinal clinical reasoning.
CURENet accurately predicts conditions that may manifest or be diagnosed in future visits, demonstrating its ability to capture patterns associated with the long-term impact of chronic illnesses and to forecast their development.
Such cases highlight the model’s predictive accuracy and interpretability, offering clinically plausible justifications that support real-time decision-making.}

\subsection{Disease Embedding Analysis (RQ5)}\label{sec:embedding}

The disease embeddings generated by LoRA Medical-Llama3-8B and our proposed CURENet model exhibit significant differences across both the MIMIC-III and FEMH datasets, as illustrated in Figure \ref{fig:vis-clustering}. CURENet consistently demonstrates more distinctive and well-delineated clusters for individual conditions across both datasets, suggesting enhanced discriminative capabilities compared with those of the baseline model. The LoRA Medical-Llama3-8B model produces embeddings with substantial cluster overlap, potentially limiting its capacity to differentiate between clinically distinct conditions. In contrast, CURENet generates embeddings with clearer boundaries between disease clusters, indicating a more nuanced representation of disease characteristics and relationships.

In the MIMIC-III dataset, both models yield relatively concentrated clusters, likely reflecting more straightforward disease correlations or dataset homogeneity. However, CURENet achieves more pronounced separation, particularly for diabetic mellitus (DM), congestive heart failure (CHF), and hypertension (HTN), demonstrating its ability to capture disease-specific features with greater fidelity. The FEMH dataset results revealed more dispersed clusters overall, presumably due to greater clinical complexity and heterogeneity in this population. Despite this increased complexity, CURENet maintains coherent disease-specific clusters, highlighting its robustness in identifying distinctive disease patterns across diverse clinical contexts.

The superior quality of CURENet's disease embeddings offers potential advantages for various clinical applications requiring precise disease differentiation, including trajectory prediction, comorbidity analysis, and personalized treatment planning. In contrast, the overlapping representations produced by LoRA Medical-Llama3-8B may limit its utility in scenarios where an accurate distinction of disease characteristics is essential, potentially resulting in overgeneralized output or increased risk of misclassification. Thus, CURENet provides a more interpretable and clinically relevant representation of disease embeddings, with distinct clusters that facilitate visualization and understanding of disease-disease relationships. Its enhanced specificity and interpretability make CURENet particularly suitable for clinical decision support systems and research applications where a comprehensive understanding of disease patterns is paramount.

\begin{figure}[h!]
    \centering
    \includegraphics[width=\textwidth]{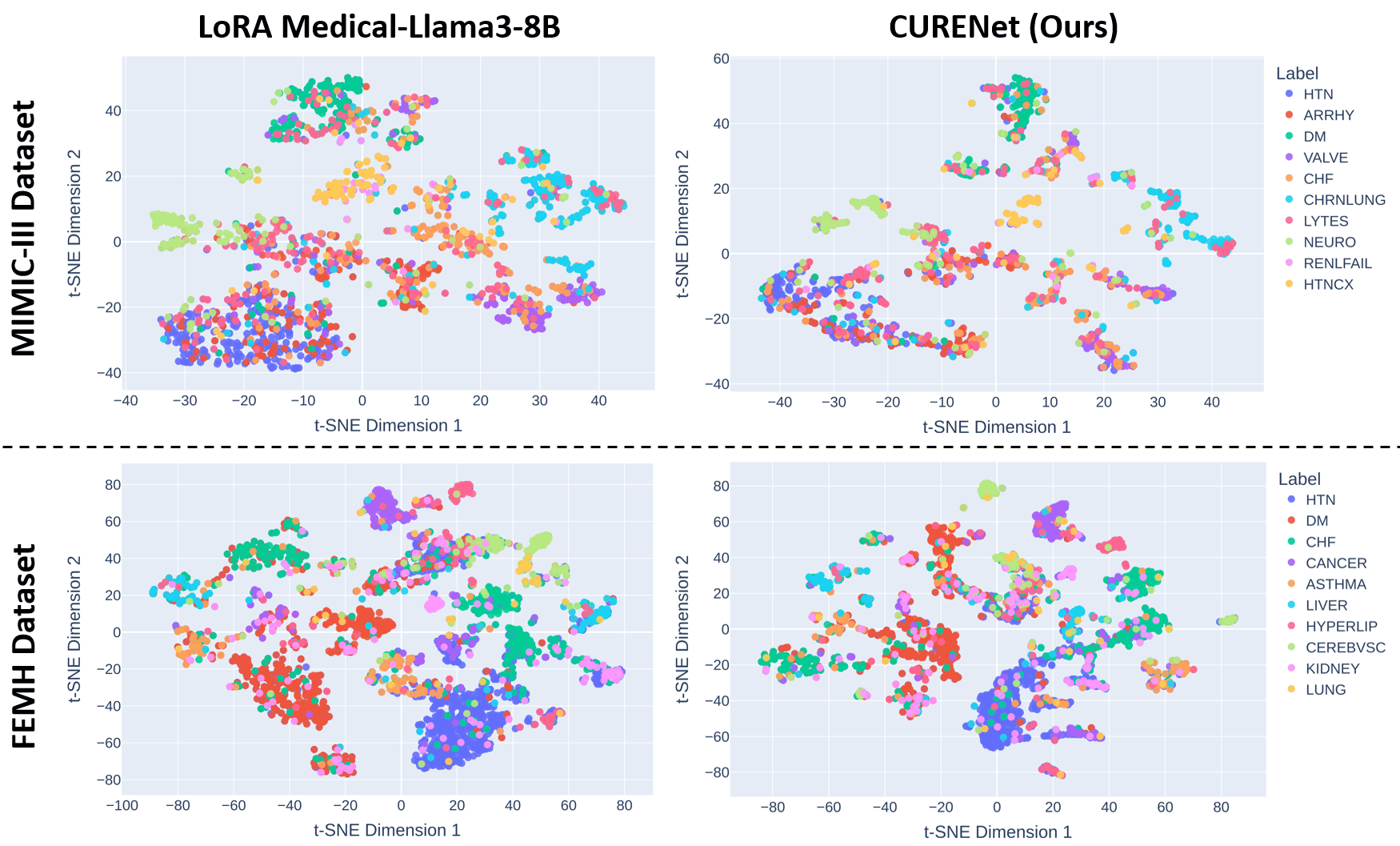}
    \caption{Disease embedding visualizations on the MIMIC-III and FEMH datasets via t-SNE obtained via LoRA Medical-Llama3-8B (left) and our proposed model CURENet (right).}
    \label{fig:vis-clustering}
\end{figure}

\section{Discussion}
\hl{CURENet effectively integrates heterogeneous EHR modalities, such as time series data, lab findings, and unstructured clinical notes, into a unified end-to-end predictive framework.
By aligning LLM-derived semantic features with transformer-based temporal modeling, CURENet addresses a key limitation of prior single-modality or shallow fusion methods, which often fail to capture cross-modal dependencies \cite{luo2020hitanet,ma2017dipole,li2020behrt,xu2018raim}.
This deep coupling between temporal and semantic features enables the discovery of clinically meaningful relationships that would otherwise remain latent.
Extensive evaluation on benchmark datasets such as MIMIC-III and FEMH shows that CURENet consistently outperforms state-of-the-art multimodal and longitudinal EHR models. The performance gains are particularly pronounced in chronic disease prediction, where the synergy between modalities enhances recall without compromising precision.}

\hl{A notable advantage of CURENet is its explicit handling of irregular visit patterns -- an often-overlooked challenge in EHR modeling -- through the encoding of visit durations and inter-visit intervals.
Two architectural innovations drive these improvements: (i) Fine-grained cross-modal representation learning that aligns semantic embeddings from unstructured clinical notes and lab texts with temporal embeddings from a Time Series Transformer, and (ii) robust handling of non-uniform clinical timelines. These design choices yield more distinctive disease embedding clusters, consistent gains of 2-4\% over state-of-the-art baselines, and enhanced interpretability. By jointly capturing temporal irregularities and semantic richness, CURENet offers a comprehensive, clinically coherent representation of patient trajectories, improving both predictive accuracy and decision support in real-world settings.
}

\hl{In addition to performance, clinical interpretability is essential for real-world applications. CURENet enhances interpretability by combining semantically rich language model embeddings with temporally aware features, enabling the identification of influential clinical notes and time-based events. Case studies (Section \ref{sec:case_study}) show that CURENet can anticipate diagnoses before formal documentation, reflecting its ability to model latent clinical trajectories aligned with physician reasoning. Analysis of disease embeddings (Section \ref{sec:embedding}) further revealed well-separated clusters, aiding the understanding of comorbidities and disease relationships.
These interpretability features enhance model transparency, support clinical trust, and facilitate integration into medical workflows, particularly in chronic disease management, where critical insights often reside in longitudinal and unstructured data.
}


\hl{In addition the their technical contributions, this work raises important ethical and privacy considerations. Multimodal EHR data, especially unstructured clinical notes, may contain sensitive information that poses re-identification risks, even in de-identified datasets. Additionally, LLMs trained on historical clinical data may inherit and amplify underlying biases, potentially affecting fairness in predictions. To address these concerns, all the data used in this study were fully de-identified and approved by the appropriate institutional review board. Future work will explore privacy-preserving techniques to further enhance transparency, fairness, and ethical deployment.}

\hl{While CURENet achieved strong performance, several considerations remain. The datasets used (MIMIC-III and FEMH) reflect specific healthcare contexts, which may limit their generalizability to other populations. Certain modalities, such as imaging and continuous physiological monitoring, were not included, and model performance depends on the quality and completeness of the EHR data. Moreover, external datasets and prospective real-world validation scenarios were not incorporated, and future studies should address these aspects to increase the robustness and applicability across diverse clinical settings.}

\hl{Future research should extend CURENet with more explicit explainability modules, such as attention heatmaps or attribution scores, to support interpretable and actionable insights.
Incorporating domain-specific priors and engaging clinical experts in the interpretation process will be critical for real-world deployment.
Additionally, concept bottleneck models in the context of emergency room revisit forecasting \cite{tseng2024predicting} offer a promising direction for enhancing interpretability and human-in-the-loop control in explainable chronic disease modeling.
Expanding to additional modalities and employing domain-adaptive pretraining can improve generalizability across diverse clinical settings. Research efforts should also focus on robustness to missing or noisy data. Generative techniques such as GANs offer the potential to improve generalizability. Finally, lightweight or explainable variants of CURENet may facilitate integration into clinical workflows, balancing performance with transparency and trust.}

\section{Conclusion}
This study presented CURENet, a diagnostic prediction model that leverages multimodal EHR data and uses transformer encoders for time series data and LLMs for text understanding. By combining unstructured clinical notes, textual lab results, and temporal visit sequences, CURENet captures complex interactions across diverse modalities that are often overlooked by conventional models. The model achieved over 94\% accuracy in predicting the top 10 chronic diseases in a multilabel classification setting, which was validated on both the publicly available MIMIC-III dataset and the private FEMH dataset.
\hl{Beyond strong predictive performance, CURENet contributes to clinical interpretability by enabling predictions to be traced back to relevant text or temporal cues, offering clinicians transparency and insight into the model’s decision process.
This makes CURENet not only accurate but also clinically actionable and trustworthy.
Future work will focus on incorporating explainable AI (XAI) techniques, such as attention-based visualization and feature attribution, to further enhance interpretability.} Additional challenges to be addressed include generalizability, handling cold-start patients, and ensuring reliable deployment in diverse healthcare environments. Enhancing the alignment between model explanations and clinical reasoning will be essential for promoting real-world adoption and supporting informed decision-making in routine care.


\section*{Acknowledgments}
\noindent We thank Dr. Hung and Professor Peng for their invaluable guidance and support and Dr. Chen for his clinical guidance during data collection at Far Eastern Memorial Hospital (FEMH). I am also sincerely thankful to my research team members, whose collaboration and hard work have significantly contributed to this work.

\section*{Funding}
\noindent The research reported in this study was partially funded by the Far Eastern Memorial Hospital (FEMH) under funding number 112086-F. Additional support was provided by the NIH NINDS with award number R21NS135482 and the NIH NIBIB with award number R21EB033455.

\section*{Declarations}
\subsection*{Conflict of interest}
The authors declare that there are no conflict of interests, and we do not have any possible conflicts of interest.





\bibliography{references}

\end{document}